\documentclass[preprint,12pt,3p]{elsarticle}

\usepackage{amsmath,amssymb,amsthm,amsfonts}
\usepackage{mathrsfs}
\usepackage{graphicx}
\usepackage{multirow}
\usepackage{multicol}
\usepackage{booktabs}
\usepackage{rotating}
\usepackage{subfig}

\usepackage{setspace}
\usepackage[printonlyused,withpage]{acronym}

\usepackage{moreverb} 

\usepackage{xspace}

\usepackage{etex}
\usepackage{tabularx}
\usepackage{caption}
\usepackage{bm} 
\usepackage{comment}
\usepackage{verbatimcopy}
\usepackage{rotating}
\usepackage{array}

\usepackage[colorinlistoftodos,prependcaption]{todonotes}

\usepackage{textcomp}
\usepackage{paralist}
\usepackage{times}
\usepackage{pgfplots}
\pgfplotsset{}
\usepackage{tikz}
\usetikzlibrary{bayesnet}
\usepackage[inline,shortlabels]{enumitem}

\usepackage{cleveref}

\newcommand\logistic{
\tikz[scale=0.05]{ %
  \begin{axis}[hide x axis, hide y axis]
  \addplot[line width=10pt,mark=none](\x, {1 / (1 + exp(-\x))}); %
  \end{axis}
}
}

\newcolumntype{C}[1]{>{\centering\let\newline\\\arraybackslash\hspace{0pt}}m{#1}}

\newcommand\Autoref[1]{\@first@ref#1,@}
\def\@throw@dot#1.#2@{#1}
\def\@set@refname#1{
    \edef\@tmp{\getrefbykeydefault{#1}{anchor}{}}%
    \def\@refname{\@nameuse{\expandafter\@throw@dot\@tmp.@autorefname}s}%
}
\def\@first@ref#1,#2{%
  \ifx#2@\autoref{#1}\let\@nextref\@gobble
  \else%
    \@set@refname{#1}
    \@refname~\ref{#1}
    \let\@nextref\@next@ref
  \fi%
  \@nextref#2%
}
\def\@next@ref#1,#2{%
   \ifx#2@ and~\ref{#1}\let\@nextref\@gobble
   \else, \ref{#1}
   \fi%
   \@nextref#2%
}

\newcommand{\softmax}{\texttt{softmax}\xspace}
\newcommand{\argmaxt}{\texttt{arg-max}\xspace}
\newcommand{\bpmfact}{\texttt{bpm}\xspace}

\newcommand{\z}{\mathbf{z}}
\newcommand{\x}{\mathbf{x}}

\newcommand{\w}{\mathbf{w}}

\newcommand{\Ncal}{\mathcal{N}}

\newcommand{\ie}{{\em i.e.~\/}}
\newcommand{\eg}{{\em e.g.~\/}}

\newcommand{\cf}{{\em c.f.~\/}}

\renewcommand{\lim}{\operatornamewithlimits{lim}}



\newcommand{\xin}{\mathbf{x}_{ \mathrm{in} }} 			
\newcommand{\xout}{\mathbf{x}_{ \mathrm{out} }}			

\newcommand{\fis}[1]{\mathrm{ne}(#1)}   	
\newcommand{\fx}[1]{ \mathbf{x}_{\mathrm{ne}(#1)} }   	

\journal{Information Fusion}

\acrodef{EPSRC}{Engineering and Physical Sciences Research Council}
\acrodef{SPHERE}{Sensor Platform for HEalthcare in Residential Environment}
\acrodef{IRC}{Inter\-disciplinary Research Collaboration}
\acrodef{CRF}{Conditional Random Field}
\acrodef{L-CRF}{Linear Chain \ac{CRF}}
\acrodef{IID}[iid]{Independent and Identically Distributed}
\acrodef{NLP}{Natural Language Processing}
\acrodef{SVM}{Support Vector Machine}
\acrodef{RBF}{Radial Basis Function}
\acrodef{RF}{Random Forest}
\acrodef{LR}{Logistic Regression}
\acrodef{POS}{Part of Speech}
\acrodef{LR}{Logistic Regression}
\acrodef{ODC}{Occasionally Dishonest Casino}
\acrodef{CASAS}{Centre for Advanced Studies in Adaptive Systems}
\acrodef{PIR}{Passive Infra-Red}
\acrodef{ADL}{Activities of Daily Living}
\acrodefplural{ADL}{Activities of Daily Living}
\acrodef{AAL}{Ambient Assisted Living}
\acrodef{MLE}{Maximum Likelihood Estimate}
\acrodef{RBF}{Radial Basis Function}
\acrodef{RKHS}{reproducing kernel Hilbert space}
\acrodef{HMM}{Hidden Markov Model}
\acrodef{AR}{Activity Recognition}
\acrodef{ERM}{Empirical Risk Minimisation}
\acrodef{AC}{Auto-correlation}
\acrodef{GP}{Gaussian Process}
\acrodef{EP}{Expectation Propagation}
\acrodef{BPM}{Bayes Point Machine}
\acrodef{MKL}{Multiple Kernel Learning}
\acrodef{MVL}{Multi-View Learning}
\acrodef{MSL}{Multi-Source Learning}
\acrodef{VMP}{Variational Message Passing}
\acrodef{EP}{Expectation Propagation}
\acrodef{RSSI}{Received Signal Strength Indication}
\acrodef{NTP}{network time protocol}
\acrodef{QoS}{Quality of Service}
\acrodef{RGB}{Red-Green-Blue}
\acrodef{RGBD}{RGB-Depth}
\acrodef{BP}{Belief Propagation}

\begin{document}

\begin{frontmatter}

\title{Probabilistic Sensor Fusion for Ambient Assisted Living}

\author[label1]{Tom Diethe\corref{cor1}}
\address[label1]{Intelligent Systems Laboratory, University of Bristol}

\cortext[cor1]{I am corresponding author}

\ead{tom.diethe@bristol.ac.uk}
\ead[url]{www.tomdiethe.com}

\author[label1]{Niall Twomey}
\ead{niall.twomey@bristol.ac.uk}

\author[label1]{Meelis Kull}
\ead{meelis.kull@bristol.ac.uk}








\author[label1]{Peter Flach}
\ead{peter.flach@bristol.ac.uk}


\author[label1]{Ian Craddock}
\ead{ian.craddock@bristol.ac.uk}

\begin{abstract}
There is a widely-accepted need to revise current forms of health-care provision, with particular interest in sensing systems in the home. Given a multiple-modality sensor platform with heterogeneous network connectivity, as is under development in the \acf{SPHERE} \acf{IRC}, we face specific challenges relating to the fusion of the heterogeneous sensor modalities.


We introduce Bayesian models for sensor fusion, which aims to address the challenges of fusion of heterogeneous sensor modalities. 
Using this approach we are able to identify the modalities that have most utility for each particular activity, and simultaneously identify which features within that activity are most relevant for a given activity. 

We further show how the two separate tasks of location prediction and activity recognition can be fused into a single model, which allows for simultaneous learning an prediction for both tasks.

We analyse the performance of this model on data collected in the \ac{SPHERE} house, and show its utility. We also compare against some benchmark models which do not have the full structure, and show how the proposed model compares favourably to these methods.
\end{abstract}


\end{frontmatter}


\section{Introduction}
\label{sec:introduction}

\subsection{Ambient and Assisted Living}
Due to well-known demographic challenges, traditional regimes of health-care are in need of re-examination. Many countries are experiencing the effects of an ageing population, which coupled with a rise in chronic health conditions is expediting a shift towards the management of a wide variety of health related issues in the home. In this context, advances in \ac{AAL} are providing resources to improve the experience of patients, as well as informing necessary interventions from relatives, carers and health-care professionals.

To this end the \acs{EPSRC}-funded ``\acf{SPHERE}'' \acf{IRC} \cite{diethe2014sphere,Woznowski15,zhu2015bridging,woznowski2016sphere} has designed a multi-modal system driven by data analytics requirements. The system is under test in a single house, and deployment to a general population of 100 homes in Bristol (UK) is underway at the time of writing. Wherever possible, the data collected will be made available to researchers in a variety of communities.

Data fusion and machine learning in this setting is required to address two main challenges: transparent decision making under uncertainty; and adapting to multiple operating contexts. Here we focus on the first of these challenges, by describing an approach to sensor fusions that takes a principled approach to the quantification of uncertainty whilst maintaining the ability to introspect on the decisions being made.

\subsection{Quantification of Uncertainty}
Multiple heterogeneous sensors in a real world environment introduce different sources of uncertainty. At a basic level, we might have sensors that are simply not working, or that are giving incorrect readings. More generally, a given sensor will at any given time have a particular signal to noise ratio, and the types of noise that are corrupting the signal might also vary. 

As a result we need to be able to handle quantities whose values are uncertain, and we need a principled framework for quantifying uncertainty which will allow us to build solutions in ways that can represent and process uncertain values. A compelling approach is to build a model of the data-generating process, which directly incorporates the noise models for each of the sensors. Probabilistic (Bayesian) graphical models, coupled with efficient inference algorithms, provide a principled and flexible modelling framework \cite{Bishop13,winn2015model}.

\subsection{Feature Construction, Selection, and Fusion}
Given an understanding of data generation processes, the sensor data can be interpreted for the identification of meaningful features. Hence it is important that this is closely coupled to the development of the individual sensing modalities \cite{Liu98}, e.g. it may be that sensors have strong spatial or temporal correlations or that specific combinations of sensors are particularly meaningful. 

One of the main hypotheses underlying the \ac{SPHERE} project \cite{diethe2014sphere} is that once calibrated, many weak signals from particular sensors can be fused into a strong signal allowing meaningful health-related interventions \cite{klein2004sensor}.

Based on the calibrated and fused signals, the system must decide whether intervention is required and which intervention to recommend; interventions will need to be information gathering as well as health providing. This is known as the exploration-exploitation dilemma, which must be extended to address the challenges of costly interventions and complex data-structures \cite{may2012optimistic}. 



Continuous streams of data can be mined for temporal patterns that vary between individuals. These temporal patterns can be directly built into the model-based framework, and additionally can be learnt on both group-wide and individual levels to learn context sensitive and specific patterns. For a recent review of methods for dealing with multiple heterogeneous streams of data in an online setting see \cite{diethe2013online}.

\section{Related Work}
\label{sec:related_work}

We consider data- or sensor- fusion in the setting of supervised learning. There are several different approaches to this, which have subtle distinctions in their motivation. In \ac{MVL}, we have multiple views of the same underlying semantic object, which may be derived from different sensors, or different sensing techniques \cite{diethe2008multiview,diethe2010constructing}. In \ac{MSL}, we have multiple sources of data which come from different sources but whose label space is aligned. Finally, in \ac{MKL} \cite{bach2004multiple}, we have multiple kernels built from different feature mappings of the same data source. In general, any algorithm built to solve any of the three problems will also solve the others. Probabilistic approaches have been developed for the \ac{MKL} problem \cite{damoulas2008probabilistic}, which gives the advantages of full distributions over the parameters, but also removes the need for heuristic methods for selecting hyperparameters. Our approach is closest in flavour to this last method.

\section{The SPHERE Challenge}
\label{sec:sphere}


In this work, we use the \ac{SPHERE} challenge dataset \cite{twomey2016sphere} as our primary source of data. This dataset contains synchronised accelerometer, environmental and video data that was recorded in a smart home by the \ac{SPHERE} project \cite{zhu2015bridging, woznowski2016sphere}. 

A number of features make this dataset valuable and interesting to activity recognition researchers and more generally to machine learning researchers: 

\begin{itemize}
    \item It features missing data. 
    \item The data are temporal, and correlations in the data must be captured either in the features or in the modelling framework \cite{twomey2016structure}. 
    \item There are many correlations between activities. 
    \item The problem can be modelled in a hierarchical classification framework.  
    \item The targets are probabilistic (since the annotations of multiple annotators were averaged). 
    \item The optimal solution will need to consider sensor fusion techniques. 
\end{itemize}

Three primary sensing modalities were collected in this dataset: 1) environmental sensor data; 2) accelerometer and \ac{RSSI} data; and 3) video and depth data. Accompanying these data are annotations on location within the smart home, as well as annotations relating to the \acp{ADL} that were being performed at the time.

\begin{figure}
    \centering
    \subfloat[Ground floor (living room, study, kitchen, downstairs hallway)]{
        \includegraphics[width=.4\textwidth]{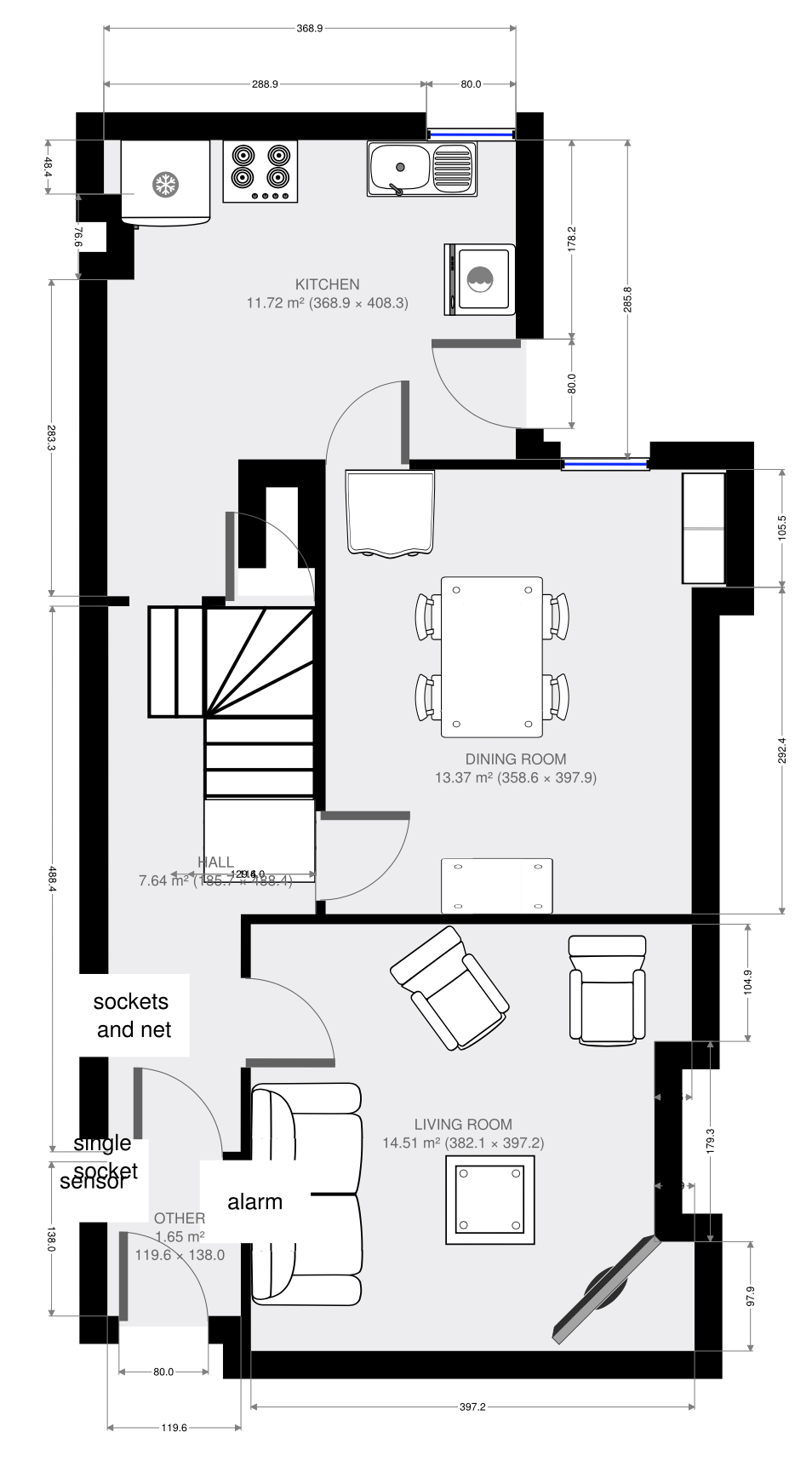}
        \label{fig:floor0}
    } ~
    \subfloat[Second floor (toilet, master bedroom, second bathroom)]{
        \includegraphics[width=.4\textwidth]{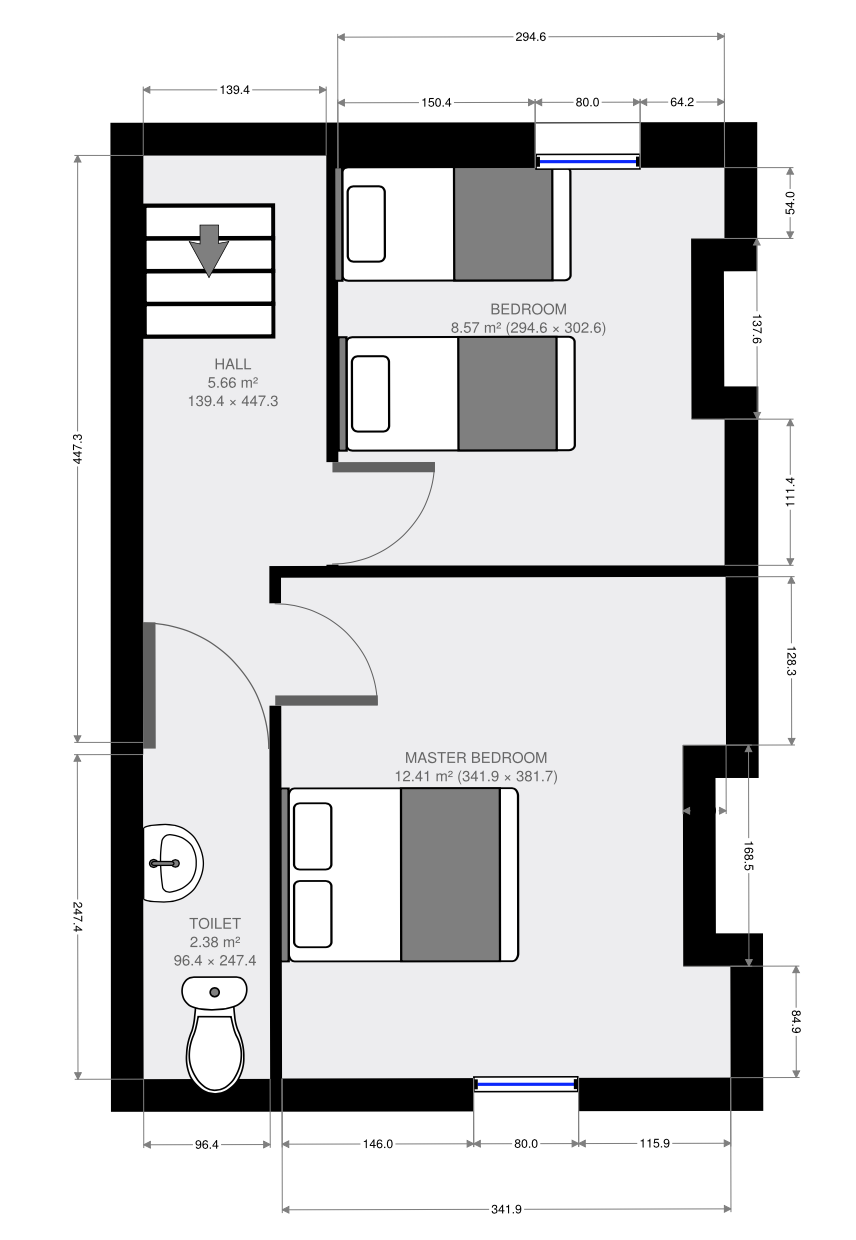}
        \label{fig:floor1}
    } //
    \subfloat[First floor (bathroom only)]{
        \includegraphics[width=.3\textwidth]{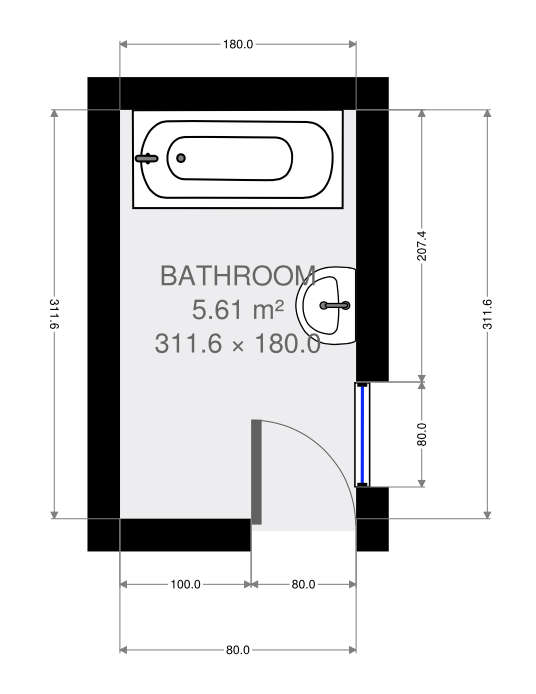}
        \label{fig:floor2}
    }
    \caption{Floor plan of the SPHERE house. A staircase joins the ground floor to the second floor, with the bathroom half-way up.} 
    \label{fig:floorplan} 
\end{figure}


The floorplan of the smart environment is shown in \autoref{fig:floorplan}, and we can see that nine rooms are marked in this figure: 

\begin{multicols}{3}
    \begin{enumerate}
        \item bathroom 
        \item bedroom 1
        \item bedroom 2
        \item hallway
        \item kitchen
        \item living room
        \item stairs
        \item study
        \item toilet
    \end{enumerate}
\end{multicols}

to predict posture and ambulation labels given the sensor data from recruited participants as they perform activities of daily living in a smart environment. 

Additionally, twenty activities of daily living are annotated in this dataset. These are enumerated below: 

\begin{multicols}{3}
    \begin{enumerate}
        \item ascent stairs;
        \item descent stairs;
        \item jump; 
        \item walk with load; 
        \item walk;
        \item bending; 
        \item kneeling; 
        \item lying; 
        \item sitting; 
        \item squatting; 
        \item standing; 
        \item stand-to-bend; 
        \item kneel-to-stand; 
        \item lie-to-sit; 
        \item sit-to-lie; 
        \item sit-to-stand;
        \item stand-to-kneel; 
        \item stand-to-sit; 
        \item bend-to-stand;  and
        \item turn
    \end{enumerate}
\end{multicols}

Three main categories of \ac{ADL} are found here: 1) ambulation activities (\ie an activity requiring of continuing movement, \eg walking), 2) static postures (\ie times when the participants are stationary, \eg standing, sitting); and 3) posture-to-posture transitions (\eg stand-to-sit, stand-to-bend). 

By considering the floorplan and the list of \acp{ADL} together, it should be clear to see that there are correlations between these: since there is no seating area in the kitchen, it is very unlikely that somebody will sit in the kitchen. However, since there are many seats in the living room, it is much more likely that somebody will sit in the living room than in the kitchen. Additionally, one can only ascent and descent stairs when one is on the staircase.

\subsection{Sensor data}
In this section we outline the individual sensor modalities in greater details. Additionally, we indicate whether each modality will be suitable for the prediction location and \acp{ADL}. 

\subsubsection{Environmental data}
\begin{figure}
  \centering
  \includegraphics[width=.95\linewidth]{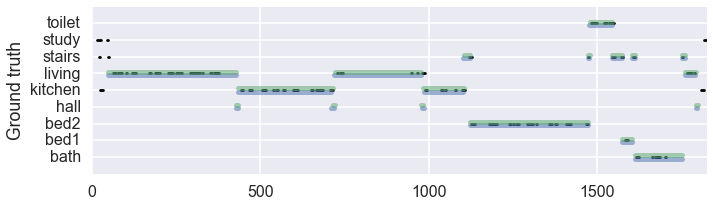}
    \caption{Example \ac{PIR} for training record \text{00001}. The black lines indicate the time durations where a \ac{PIR} is activated. The blue and green wide horizontal lines indicate the room occupancy labels as given by the two annotators that labelled this sequence.}
    \label{fig:example_pir}
\end{figure}

In \autoref{fig:example_pir}, we show the \ac{PIR} data that was recorded and localisation labels that were annotated as part of the data collection. 
In this image, the black segments indicate the times during which motion was detected by the \ac{PIR} sensor. The remaining line segments (\eg the transparent blue and green segments) represent the ground truth location annotations. In this example, two colours (blue and green) are shown since two annotators labelled this segment. Overall, these localisation annotations are very similar overall, and the main differences are on the precise specification of `start' and `end' times in each room, see for example the `hall' annotation at approximately $1\,000$ seconds. 

The \ac{PIR} data is suitable only for prediction of localisation since it cannot report on fine-grained movements. However, this sensor provides useful information pertaining to localisation since each of the nine rooms contains these sensors. 

It is worth noting that under certain circumstances the \ac{PIR} data may report `false positive' triggers, \eg when bright sunshine is present. For this reason, it is necessary to fuse additional modalities together to produce accurate localisation.

\subsubsection{Accelerometer and \ac{RSSI} data}

\begin{figure}
    \centering
    \subfloat[Acceleration signal trace shown over a 5 minute time period. Annotations are overlaid in blue and green.]{
        \includegraphics[width=0.7\linewidth]{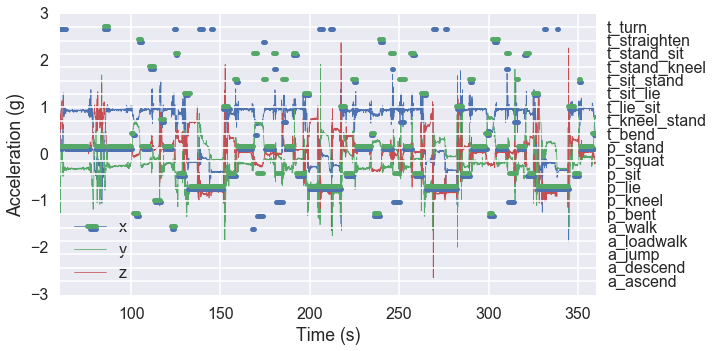}
        \label{fig:accel}
    } //
    \subfloat[\ac{RSSI} values from the four access points. Room occupancy labels are shown by the the horizontal lines.]{
        \includegraphics[width=0.7\linewidth]{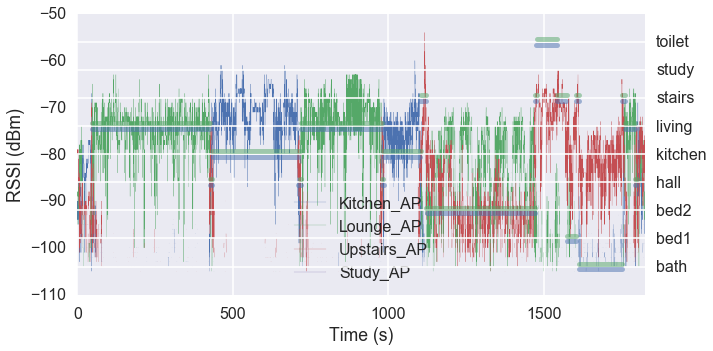}
        \label{fig:rssi}
    }
    \caption{Example acceleration and \ac{RSSI} signals for training record \text{00001}. The line traces indicate the accelerometer/\ac{RSSI} values recorded by the access points. The horizontal lines indicate the ground-truth as provided by the annotators (two annotators annotated this record, and their annotations are depicted by the green and blue traces respectively.} 
    \label{fig:acceleration_rssi} 
\end{figure}

The accelerometer signal trace and the annotated \acp{ADL} are shown in \autoref{fig:accel}. The continuous blue, green and red traces here are related to the $x$, $y$, and $z$ axes of the accelerometer. These signals are assigned to the left hand axis. 

%
Overlaid on this image are discontinuous line segments in blue and green, and these are the \ac{ADL} annotations that were produced from two annotators. One should confer with the right hand axis to identify the label at a particular time point. 

In general, there is good agreement between these two annotators. However, some annotators naturally produced greater resolution than others (\eg the `blue' annotator annotated more turning activities than the `green' annotator). 

Since the accelerometer data are transmitted to central access points via a wireless Bluetooth connection, we also have access to the relative strength of the wireless connection at each of the access points. Four access points were distributed throughout the smart home: one each in the kitchen, living room, master bedroom, and the study. The \ac{RSSI} signals that were recorded by the four access points are shown in \autoref{fig:rssi}, and the localisation labels are also overliad on this image (\cf to \autoref{fig:example_pir}). 

This image shows that there is a close relationship between the values of \ac{RSSI} and location. Taking a concrete example, we can see that during the first 500 seconds, a significant proportion of time is spent in the living room. The access point in the living room (\ie the green trace) reports the highest response, and the remaining access points have not registered the connection. Consequently, the \ac{RSSI} and the \ac{PIR} are both valuable signals for localisation. 

It is worth noting that the Bluetooth link is set up in `connectionless' mode. This mode increases the live cycle of the wearable battery, and also allows the wearable data to be acquired simultaneously by multiple access points. However, the data are also transmitted with a lower \ac{QoS} which increases the risk of lost packets. This is particularly clear in \autoref{fig:rssi} since at no time is data received on all access points. Similarly, in \autoref{fig:accel} acceleration packets are lost.

\subsubsection{Video data}

Video recordings were taken using ASUS Xtion PRO \ac{RGBD} cameras\footnote{\url{https://www.asus.com/3D-Sensor/Xtion_PRO/}}. Automatic detection of humans was performed using the OpenNI library\footnote{\url{https://github.com/OpenNI/OpenNI}}. False positive detections were manually removed by the organisers by visual inspection. Three \ac{RGBD} cameras are installed in the \ac{SPHERE} house, and these are located in the living room, downstairs hallway, and the kitchen. No cameras are located elsewhere in the residence. 

In order to preserve the anonymity of the participants the raw video data are not available, and instead the coordinates of the 2D bounding box, 2D centre of mass, 3D bounding box and 3D centre of mass are provided. The units of 2D coordinates are in pixels (i.e. number of pixels down and right from the upper left hand corner) from an image of size $640\times480$ pixels. The coordinate system of the 3D data is axis aligned with the 2D bounding box, with a supplementary dimension that projects from the central position of the video frames. The first two dimensions specify the vertical and horizontal displacement of a point from the central vector (in millimetres), and the final dimension specifies the projection of the object along the central vector (again, in millimetres). 

\begin{figure}[t]
  \centering
  \includegraphics[width=.8\linewidth]{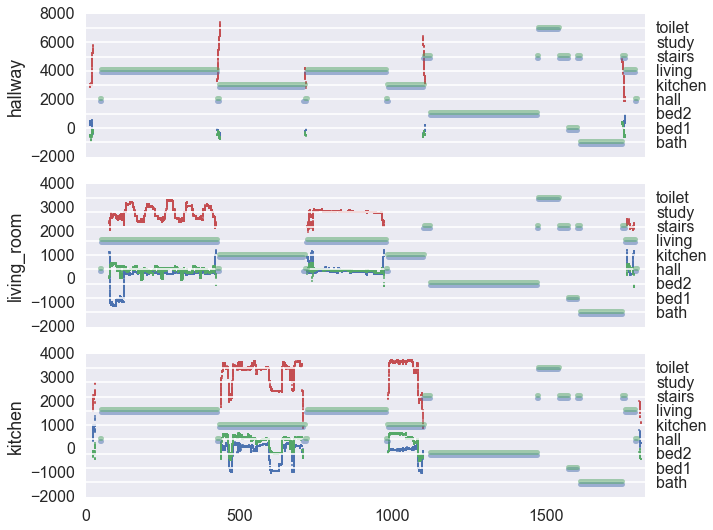}
    \caption{Example \texttt{centre\_3d} for training record \text{00001}. The horizontal lines indicate annotated room occupancy. The blue, green and red traces are the $x$, $y$, and $z$ values for the 3D centre of mass.}
    \label{fig:example_video}
\end{figure}

Examples of the centre of the 3D bounding box are shown in \autoref{fig:example_video} where the three video locations are separated according to their location. Video data is rich in the information that it provides, and it is informative for predicting location and \acp{ADL}. Room-level location can be inferred since each camera is situated in a room. Additionally, with appropriate feature engineering, it is possible to determine activities of daily living. However, since cameras are only available in three rooms, we cannot rely purely on \ac{RGBD} features for localisation or \ac{ADL}.

\subsection{Recruitment and Sensor Layout}
Twenty participants were recruited to perform activities of daily living in the \ac{SPHERE} house while the data recorded by the sensors was logged to a database.  Ethical approval was secured from the University of Bristol's ethics committee to conduct data collection, and informed consent was obtained from healthy volunteers. 

In the first stage of data collection, participants were requested to follow a pre-defined script. \autoref{fig:floor0} \autoref{fig:floor1} and \autoref{fig:floor2} show the floor plan of the ground, first and second floors of the smart environment respectively.

\subsection{Annotation}\label{section:annotation}
A team of 12 annotators were recruited and trained to annotate the locations of \acp{ADL}. 
To support the annotation process, a head mounted camera (Panasonic HX-A500E-K 4K Wearable Action Camera Camcorder) recorded 4K video at 25 FPS to an SD-card. This data is not shared in this dataset, and is used only to assist the annotators. Synchronisation between the \ac{NTP} clock and the head-mounted camera was achieved by focusing the camera on an \ac{NTP}-synchronised digital clock at the beginning and end of the recording sequences. 

An annotation tool called ELAN\footnote{\url{https://tla.mpi.nl/tools/tla-tools/elan/}} was used for annotation. ELAN is a tool for the creation of complex annotations on video and audio resources, developed by the Max Planck Institute for Psycholinguistics in Nijmegen, The Netherlands.

Room occupancy labels are also given in the training sequences. While performance evaluation is not directly affected by room prediction on this, participants may find that modelling room occupancy may be informative for prediction of posture and ambulation.

\subsection{Performance evaluation}
As both the targets and predictions are probabilistic, the standard classification performance evaluation metrics are inappropriate. Hence, classification performance is evaluated with weighted Brier score \cite{brier1950verification}:

\begin{align}
    BS = \frac{1}{N} \sum_{n=1}^N \sum_{c=1}^C w_c \left( p_{n, c} - y_{n, c} \right)^2
\end{align}

\noindent 
where $N$ is the number of test sequences, $C$ is the number of classes, $w_c$ is the weight for each class, $p_{n, c}$ is the predicted probability of instance $n$ being from class $c$, and $y_{n, c}$ is the proportion of annotators that labelled instance $n$ as arising from class $c$. Lower Brier score values indicate better performance, with optimal performance achieved with a Brier score of $0$.


\section{Bayesian Sensor Fusion}
\label{sec:fusion}
In this section we will develop a class of models that may be used to tackle the two sensor fusion tasks described above. Critically, we will also show how these two can be combined into a single model. 

\subsection{Preliminaries}
We will begin by assuming that distributions (or densities) over a set of variables $\x = (x_1, \ldots x_D)$ of interest can be represented as factor graphs, \ie
\begin{align}
p(\x) = \frac{1}{Z} \prod_{j=1}^J \psi_j(\fx{\psi_j}),
\end{align}
where $\psi_j$ are the \emph{factors}: non-negative functions defined over subsets of the variables $\fx{\psi_j}$, the neighbours of the factor node $\psi_j$ in the graph, where we use $\fis{\psi_j}$ to denote the set indices of the variables of factor $\psi_j$). $Z$ is the normalisation constant. We only consider here \emph{directed} factors $\psi(\xout | \xin)$ which specify a conditional distribution over variables $\xout$ given $\xin$ (hence $\fx{\psi} = (\xout, \xin))$. For more details see \eg \cite{kschischang2001factor,Bishop06}.

\subsection{Multi-Class \acl{BPM}}
We begin by describing a Bayesian model for classification known as the multi-class \ac{BPM} \cite{Herbrich01}, which will form the basis of our modelling approach. The \ac{BPM} makes the following assumptions:

\begin{enumerate}[noitemsep]
 \item The feature values $\x$ are always fully observed.
 \item The order of instances does not matter. \label{ass:order}
 \item The predictive distribution is a linear discriminant of the form $p(y_i|\x_i,\w) = p(y_i|s_i = \w'\x_i)$ where $\w$ are the weights and $s_i$ is the score for instance $i$.
 \item The scores are subject to additive Gaussian noise. \label{ass:noise}
 \item The features are as uncorrelated as possible. This allows modelling $p(\w)$ as fully factorised
 \item \emph{(optional)} A factorised heavy-tailed prior distributions over the weights $\w$
\end{enumerate}

For the purposes of both location prediction and activity recognition, assumption \ref{ass:order} may be problematic, since the data is clearly sequential in nature. Intuitively, we might imagine that the strength of the temporal dependence in the sequence will determine how costly this approximation is, and this will in turn depend on how the data is preprocessed (\ie is raw data presented to the classifier, or are features instead computed from the time series?). It has been shown \cite{twomey2016structure} that under certain conditions structured models and unstructured models can yield equivalent predictive performance on sequential tasks, whilst unstructured models are also typically much cheaper to compute. We follow the guidance set out there and use contextual information from neighbouring time-points to construct our features.

\begin{figure}[ht]
    \centering
    \subfloat[]{
        \includegraphics[width=0.3\textwidth]{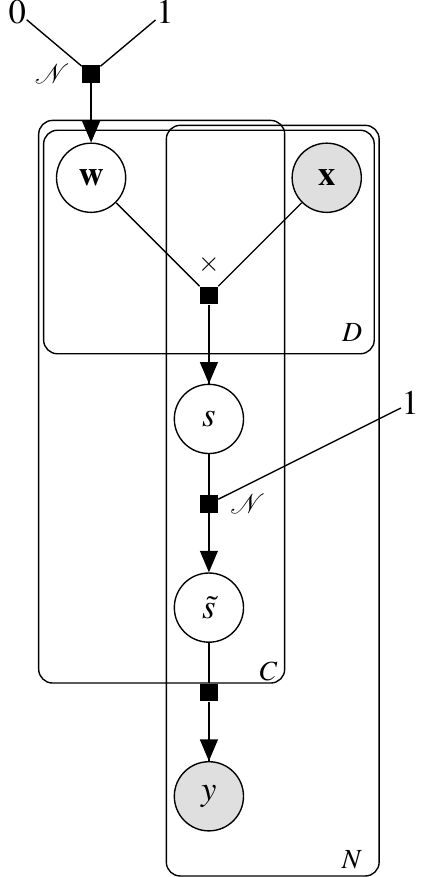}
        \label{fig:bpm}
    } ~
    \subfloat[]{
        \includegraphics[width=0.3\textwidth]{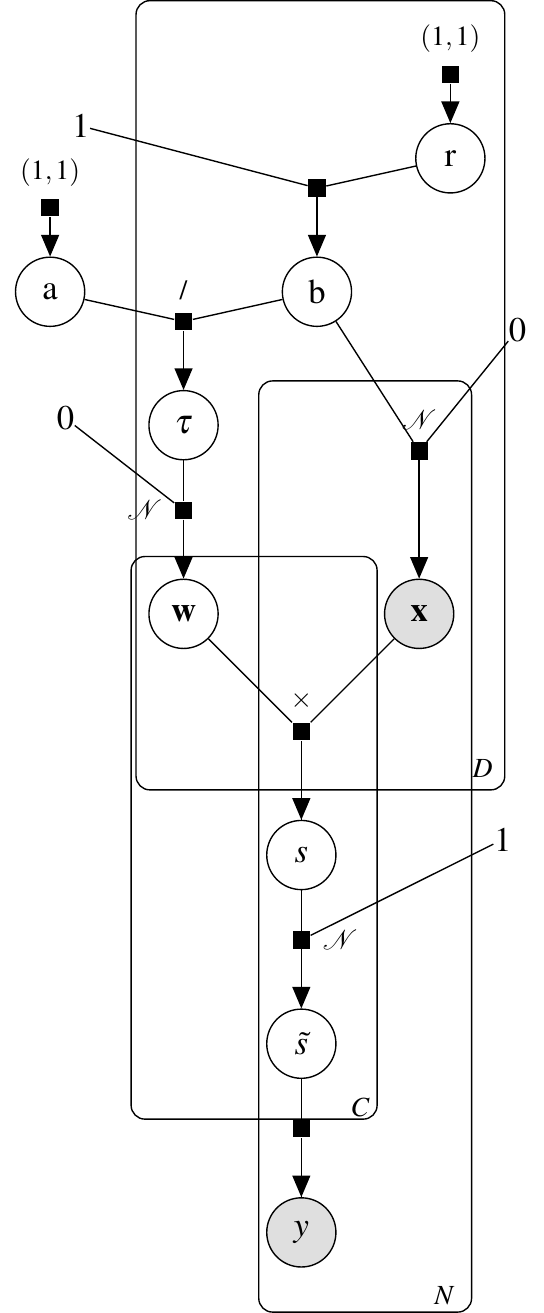}
    \label{fig:heavy_tails}
    } ~
    \subfloat[]{
        \includegraphics[width=0.2\textwidth]{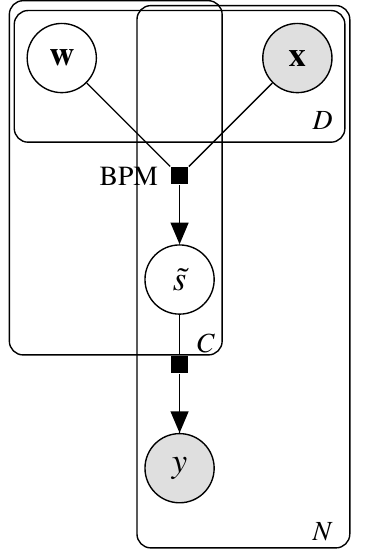}
    \label{fig:simplified}
    }
    \caption{(a) Multi-class \acl{BPM}; (b) heavy-tailed version; and (c) simplified representation. $C$: Number of classes; $N$: Number of examples; $D$: Number of features.  See text for details.} 
\end{figure}

The factor graph for the basic multi-class \ac{BPM} model is illustrated in \autoref{fig:bpm}, where $\Ncal$ denotes a Gaussian density for a given mean $\mu$ and precision $\tau$, and $\Gamma$ denotes a Gamma density for given shape $k$ and scale $\theta$. The factor indicated by $\logistic$ is the \argmaxt factor, which is like a probabilistic multi-class switch. The additive Gaussian noise from assumption \ref{ass:noise} results in the variable $\tilde{s}$, which is a noisy version of the score $s$.

An alternative to the \argmaxt factor is the \softmax function, or normalised exponential function, which is a generalisation of the logistic function that transforms a $K$-dimensional vector $\z$  of arbitrary real values to a $K$-dimensional vector $\sigma(\z)$ of real values in the range $(0, 1)$ that sum to 1. The function is given by
\begin{align*}
\sigma(\z_{j})= \frac {e^{\z_j}}{\sum_{k=1}^{K} e^{\z_k}} \quad \quad \quad j = 1, \ldots, K.
\end{align*}
Multi-class regression involves constructing a separate linear regression for each class, the output of which is known as an auxiliary variable. The vector of auxiliary variables for all classes is put through the \softmax function to give a probability vector, corresponding to the probability of being in each class. 

Multi-class \softmax regression is very similar in spirit to the multi-class \ac{BPM}. The key differences are that where the \ac{BPM} uses a max operation constructed from multiple ``greater than'' factors, \softmax regression uses the \softmax factor, and \ac{VMP} (see \autoref{sec:inference}) is currently the only option when using the \softmax factor. Two consequences of these differences are that \softmax regression scales computationally much better with the number of classes than the \ac{BPM} (linear rather quadratic complexity) and although not relevant here it is possible to have multiple counts for a single sample (multinomial regression). In this case the number of classes $C$ is not so large, meaning that the quadratic $O(C^2)$ scaling of the \ac{BPM} is not so problematic.

We also propose to use a heavy-tailed prior, a Gaussian distribution whose precision is itself a mixture of Gamma-Gamma distributions. This is illustrated in the factor graph in \autoref{fig:heavy_tails}, where $\Ncal$ denotes a Gaussian density for a given mean $\mu$ and precision $\tau$, and $\Gamma$ denotes a Gamma density for given shape and rate). The variable $a$ is a common precision that is shared by all features, and the variable $b$ represents the precision rate, which adapts to the scaling of the features. Compared with a Gaussian prior, the heavy-tailed prior is more robust towards outliers, \ie feature values which are far from the mean of the weight distribution, 0. It is invariant to rescaling the feature values, but not invariant to their shifting, which is achieved by adding a constant feature value (bias) for all instances. 

In \autoref{fig:simplified} we also give a simplified form of the factor graph, where the essential parts of the \ac{BPM} have been collapsed into a single \bpmfact factor. We will use the simplified form in further models to aid the presentation, but should be born in mind that this actually represents \autoref{fig:bpm} or \autoref{fig:heavy_tails} depending on the context.

In this form, the model is not identifiable, since the addition of a constant value to all the score variables $s$ will not change the output of the prediction. To make the model identifiable we enforce that the score variable corresponding to the last class will be zero, by constraining the last coefficient vector and mean to be zero. 

\subsection{Fused \acl{BPM}}
The simplest method for fusing multiple sensing modalities is to concatenate all of the features from each modality for each example into a single long feature vector. This can then be presented to the \ac{BPM} as described above, and will serve as a baseline model. 

\begin{figure}[ht]
    \centering
    \subfloat[]{
        \includegraphics{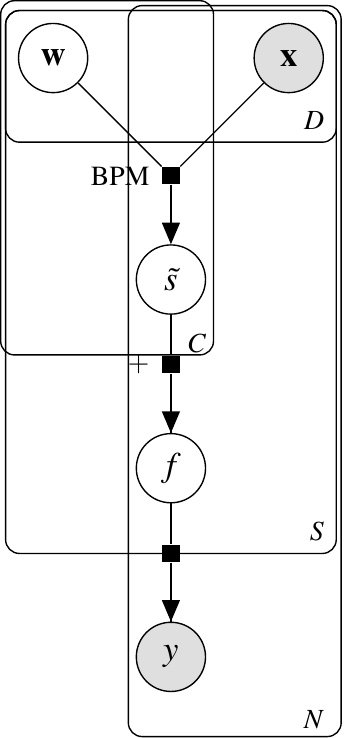}
    \label{fig:fusion_add}
    } ~
    \subfloat[]{
        \includegraphics{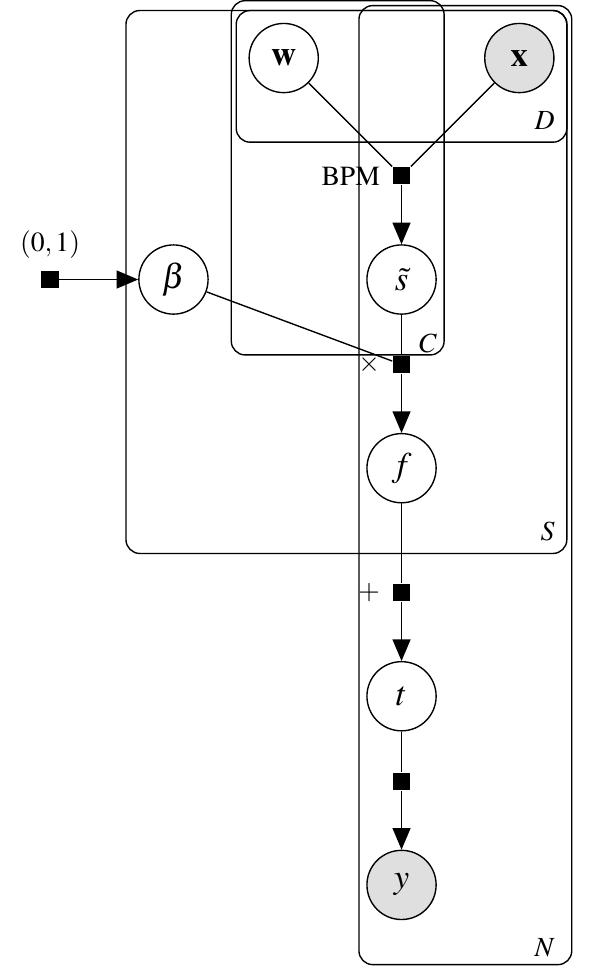}
    \label{fig:fused_weighted}
    } ~
    \subfloat[]{
        \includegraphics{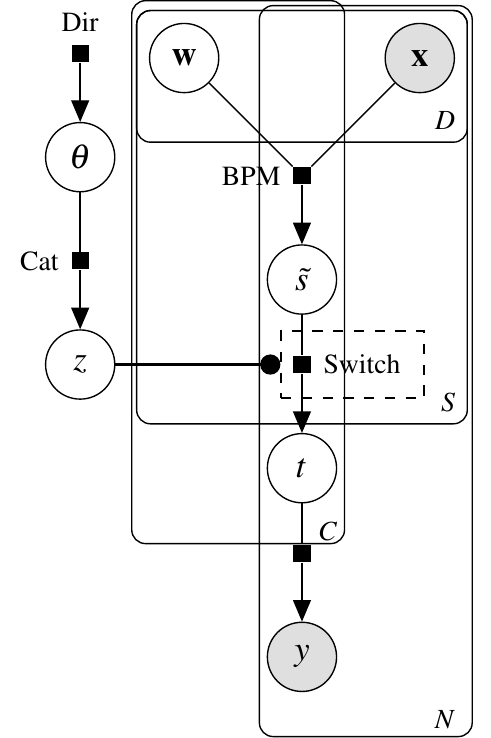}
    \label{fig:fused_switch}
    }
    \caption{Bayesian multi-class fusion classification models: (a) unweighted additive version; (b) weighted additive version; and (c) fused switching model. $S$: Number of sources. See text for details.} 
\label{fig:fusion}
\end{figure}

Here we present three Bayesian models for sensor fusion. the first of these, shown in \autoref{fig:fusion_add}, is the simplest of these, and corresponds quite closesly to the unweighted probabilistic \ac{MKL} model given in \cite{damoulas2008probabilistic}. Taking the simplified factor graph in \autoref{fig:simplified}, we have an additional plate over the sources/modalities $S$, for the entire \ac{BPM} model, which results in a noisy score $\tilde{s}$ for each source. These are then summed together to get a fused score per class, and the standard \argmaxt or \softmax link function can then be used.

The first modification of this is the weighted additive fusion model shown in \autoref{fig:fused_weighted}. Here we have additional variables $\beta$ per source, which are than multiplied with the noisy scores before addition to give fused scores. This model provides extra interpretability since the means of the $\beta$ variables indicate the influence of the source associated with that variable, and the variances of these variables indicate the overall uncertainty about that source. However this model also introduces extra symmetries that can cause difficulties in inference. 

The second modification is the switching model presented in \autoref{fig:fused_switch}. In this model, we have a single switching variable $z$ which follows a categorical distribution that is defined over the range of the sources. This categorical distribution is parametrised by the probability vector $\theta$, which follows a uniform Dirichlet distribution.

\subsection{Stacked classifiers}
The models described thus far have all been designed to tackle fusion of multiple sources to achieve a single multi-class classification task. However, in the \ac{SPHERE} challenge setting, there are in fact two classification tasks: firstly, there is the classification of location, based on the fusion of \ac{RSSI} values from the wearable device along with the stationary \ac{PIR} sensors. Secondly, there is the classification of the activity being performed, based on the fusion of the wearable accelerometer readings and the video bounding boxes. It is a reasonable hypothesis that certain activities are much more likely to occur in some locations than others: hence, we propose another form of fusion, to train activity recognition models for each location use the location prediction model to switch over these models. 

\begin{figure}[ht]
  \centering
  \resizebox{0.2\linewidth}{!}{%
        \includegraphics{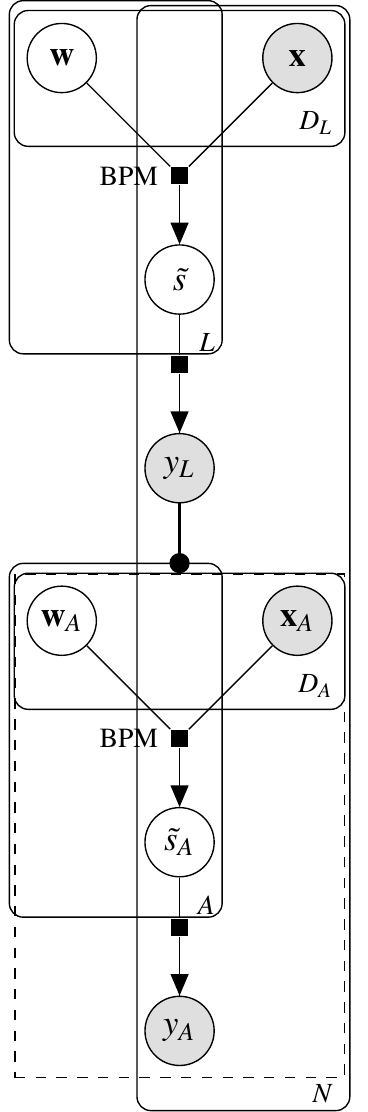}
    }
    \caption{Stacked classification model for simultaneous location prediction and activity recognition. $L$: Number of locations; $A$: Number of activities $N$: Number of examples; $D_L$: Number of location features. $D_A$: Number of activity features. See text for details.}
    \label{fig:stacked}
\end{figure}

This could of course be done in a two stage process, where the switch could take the a weighted combination of the predictions from the location predictions to produce a fused prediction (again this can form a strong baseline). However we also surmise that there may be an advantage to be gained by learning both stages in a single model, since the activity recognition models could inform the location prediction as well. This model is depicted in \autoref{fig:stacked}. Note that the location predictions $y_L$ now act as a probabilistic switching variable over the activity \acp{BPM}, of which there is one per location. For simplicity, we have omitted the fusion parts of this model, but note that we can incorporate any of the 3 kinds of fusion found in \autoref{fig:fusion}.

\section{Methodology}
\label{sec:methods}

\subsection{Inference}
\label{sec:inference}
The \ac{BP} (or sum-product) algorithm computes marginal distributions over subsets of variables by iteratively passing messages between variables and factors within a given graphical model, ensuring consistency of the obtained marginals at convergence. Unfortunately, for all but trivial models, exact inference using \ac{BP} is intractable. In this work, we employ two separate inference algorithms, both of which are efficient deterministic approximations based on message-passing over the factor graphs we have seen before. 

\ac{EP} \cite{minka2001expectation} introduces an approximation in the case when the messages from factors to variables do not have a simple parametric form, by projecting the exact marginal onto a member of some class of known parametric distributions, typically in the exponential family, \eg the set of Gaussian or Beta distributions.

\ac{VMP} applies variational inference to Bayesian graphical models \cite{winn2005variational}. \ac{VMP} proceeds by sending messages between nodes in the network and updating posterior beliefs using local operations at each node. Each such update increases a lower bound on the log evidence at that node, (unless already at a local maximum). \ac{VMP} can be applied to a very general class of conjugate-exponential models because it uses a factorised variational approximation, and by introducing additional variational parameters, \ac{VMP} can be applied to models containing non-conjugate distributions. 

\ac{EP} and \ac{VMP} has a major advantage over  sampling methods in this setting, which is that it is relatively easy to compute model evidence (see Equation \eqref{eq:evidence} in below). In the case of \ac{VMP}, this is directly available from the lower bound. The evidence computations can be used both for model comparison and for detection of convergence. 

\subsection{Model Comparison}
We would also like perform Bayesian model comparison, in which we marginalise over the parameters for the type of model being used, with the remaining variable being the identity of the model itself. The resulting marginalised likelihood, known as the model evidence, is the probability of the data given the model type, not assuming any particular model parameters. Using $D$ for data, $\theta$ to denote model parameters, $H$ as the hypothesis, the marginal likelihood for the model $H$ is
\begin{align} \label{eq:evidence}
    p(D|H) = \int p(D|\theta, H) \, p(\theta|H) \, \operatorname{d}\!\theta 
\end{align}
This quantity can then be used to compute the \emph{Bayes factor} \cite{Goodman99}, which is the posterior odds ratio for a model $H_1$ against another model $H_2$,
\begin{align}
 \frac{p(H_1|D)}{p(H_2|D)} = \frac{p(H_1)p(D|H_1)}{p(H_2)p(D|H_2)}.
\end{align}

\subsection{Batch and Online Learning}
There may be occasions when the full model will not fit in memory at once (either at train time or prediction time), as is the case for the \ac{SPHERE} challenge dataset when performing inference on a standard desktop computer. In this case, there are two options: batched inference and online learning. Infer.NET provides a \emph{SharedVariable} class, supports sharing between models of different structures and also across multiple data batches, and makes it easy to implement batched inference. Online learning is performed using the standard assumed-density filtering method of \cite{Opper98}, which in the case of the weight variables which are marginally Gaussian distributed, simply equates to setting the priors to be the poteriors from the previous round of inference.

\subsection{Data pre-processing}
We further perform standardisation (whitening) for each feature within each source independently. Whilst this is not strictly necessary for the heavy-tailed models, it will not negatively impact on their performance so is done for all models to ensure consistency. We further add a constant bias feature (equal to 1) per source.

We also augment the feature space by adding polynomial degree 2 interaction features for each source, meaning that we in effect performing polynomial regression. This is equivalent to the implicit feature space of a degree 2 polynomial kernel in kernel methods, which intuitively, means that both features and pairwise combinations of features are taken into account. When the input features are binary-valued, then the features correspond to logical conjunctions of input features. Note that this method roughly doubles the number of variables in the model, but in this case the feature spaces are of relatively low dimension.

\section{Discussion}
\label{sec:discussion}
Prediction performance is the one clear motivation for the use of models for fusion. However, even if prediction performance is not improved (or degraded in non-significant ways) it can provide other benefits. These include extra interpretability of the models by looking at how the modalities are used by the models, and potentially robustness to missing modalities, since the models can be encouraged to use all of the modalities even when they provide less information for predictions. It might be possible to achieve such robustness with a simple \ac{BPM} through re-weighting features on the basis of the posteriors from training, but it would clearly be more satisfying if this could achieved only by changing mixing coefficients to get the same effect.

\subsection{Early versus late fusion}
Generally speaking, the literature describes \emph{early} and \emph{late} fusion strategies \cite{snoek2005early}. Early fusion is a usually described as a scheme that integrates unimodal features before learning concepts. This would include strategies such as concatenating feature spaces and (probabilistic) \ac{MKL} \cite{bach2004multiple,damoulas2008probabilistic}. Late fusion is a scheme that first reduces unimodal features to separately inferred concept scores (\eg through independent classifiers), after which these scores are integrated to learn concepts, typically using a further classification algorithm. \citep{snoek2005early} showed empirically that late fusion tended to give slightly better performance for most concepts, but for those concepts where early fusion performs better the difference was more significant.

The methods described in this paper fit more in line with the late fusion scheme under this definition, since the fusion is occurring at the score level. However the framework laid out here is more principled, since there is no recourse to two sets of classification algorithms and the heuristic manipulations that they entail.

\subsection{Transfer Learning}
As shown in \cite{diethe2016active}, it is possible to extend the multi-class \ac{BPM} \cite{Herbrich01} to the transfer learning setting by adding an additional layer of hierarchy to the model. To deal with the transfer of learning between individuals (residents), we would have an extra plate around the individuals that are present in the training set ($R$), who form the \emph{community}, allowing us to learning shared weights for the community.

To apply our learnt community weight posteriors to a new individual we can use the same model configured for a single individual with the priors over weight mean and weight precision replaced by the Gaussian and Gamma posteriors learnt from the individuals in the training set respectively. This model is able to make predictions even when we have not seen any data for the new individual, but it is also possible to do online training as we receive labelled data for the individual. By doing so, we can smoothly evolve from making generic predictions that may apply to any individual to making personalised predictions specific to the new individual. Note that as stated this only applies to the version of the model without heavy-tailed priors, but can be extended to this setting as well.

Note that it is also possible to separate the community priors into groups (\eg by demographics or medical condition) by having hyper-priors for each group, a separate indicator variable indicating group membership, acting as a gating variable to select the appropriate hyper-priors. 

The separate transfer learning problem from house-to-house is achieved through the method of introducing meta-features of \cite{Rashidi11}, and then the feature space is automatically mapped from the source domain to the target domain. In order to apply this to the models discussed in \autoref{sec:fusion}, we would assume that the features $\x$ have already been mapped to these meta-features, and that similarly for the personalisation phase the mapping has already taken place.

\section{Conclusions}
\label{sec:conclusions}
There is a widely-accepted need to revise current forms of health-care provision, with particular interest in sensing systems in the home. Given a multiple-modality sensor platform with heterogeneous network connectivity, as is under development in the \acf{SPHERE} \acf{IRC}, we face specific challenges relating to the fusion of the heterogeneous sensor modalities. We further will require the transfer of learnt models to a deployment context that may differ from the training context. 


We introduce Bayesian models for sensor fusion, which aim to address the challenges of fusion of heterogeneous sensor modalities.
Using this approach we are able to identify the modalities that have most utility for each particular activity, and simultaneously identify which features within that activity are most relevant for a given activity. 

We further show how the two separate tasks of location prediction and activity recognition can be fused into a single model, which allows for simultaneous learning an prediction for both tasks.

We analyse the performance of this model on data collected in the \ac{SPHERE} house, and show its utility. We also compare against some benchmark models which do not have the full structure, and show how the proposed model compares favourably to these methods.



\subsubsection*{Acknowledgements}
This work was performed under the \ac{SPHERE} \ac{IRC} funded by the UK \ac{EPSRC}, Grant EP/K031910/1.
The project is actively working towards releasing high-quality data sets to encourage community participation in tackling the issues outlined here.



\bibliographystyle{elsarticle-harv}

\begin{thebibliography}{27}
\expandafter\ifx\csname natexlab\endcsname\relax\def\natexlab#1{#1}\fi
\expandafter\ifx\csname url\endcsname\relax
  \def\url#1{\texttt{#1}}\fi
\expandafter\ifx\csname urlprefix\endcsname\relax\def\urlprefix{URL }\fi

\bibitem[{Bach et~al.(2004)Bach, Lanckriet, and Jordan}]{bach2004multiple}
Bach, F.~R., Lanckriet, G.~R., Jordan, M.~I., 2004. Multiple kernel learning,
  conic duality, and the smo algorithm. In: Proceedings of the twenty-first
  international conference on Machine learning. ACM, p.~6.

\bibitem[{Bishop(2013)}]{Bishop13}
Bishop, C., 2013. Model-based machine learning. Phil Trans R Soc A 371.

\bibitem[{Bishop et~al.(2006)}]{Bishop06}
Bishop, C.~M., et~al., 2006. Pattern recognition and machine learning. Vol.~4.
  springer New York.

\bibitem[{Brier(1950)}]{brier1950verification}
Brier, G.~W., 1950. Verification of forecasts expressed in terms of
  probability. Monthly weather review 78~(1), 1--3.

\bibitem[{Damoulas and Girolami(2008)}]{damoulas2008probabilistic}
Damoulas, T., Girolami, M.~A., 2008. Probabilistic multi-class multi-kernel
  learning: on protein fold recognition and remote homology detection.
  Bioinformatics 24~(10), 1264--1270.

\bibitem[{Diethe and Girolami(2013)}]{diethe2013online}
Diethe, T., Girolami, M., 2013. Online learning with (multiple) kernels: A
  review. Neural Computation 25, 567--625.

\bibitem[{Diethe et~al.(2008)Diethe, Hardoon, and
  Shawe-Taylor}]{diethe2008multiview}
Diethe, T., Hardoon, D.~R., Shawe-Taylor, J., 2008. Multiview {F}isher
  discriminant analysis. In: {NIPS} 2008 workshop ``Learning from Multiple
  Sources''.

\bibitem[{Diethe et~al.(2010)Diethe, Hardoon, and
  Shawe-Taylor}]{diethe2010constructing}
Diethe, T., Hardoon, D.~R., Shawe-Taylor, J., 2010. Constructing nonlinear
  discriminants from multiple data views. In: ECML/PKDD. Vol.~1. pp. 328--343.

\bibitem[{Diethe et~al.(2014)Diethe, Twomey, and Flach}]{diethe2014sphere}
Diethe, T., Twomey, N., Flach, P., 2014. {SPHERE}: A sensor platform for
  healthcare in a residential environment. In: Proceedings of Large-scale
  Online Learning and Decision Making Workshop.

\bibitem[{Diethe et~al.(2016)Diethe, Twomey, and Flach}]{diethe2016active}
Diethe, T., Twomey, N., Flach, P., 2016. Active transfer learning for activity
  recognition. In: European Symposium on Artificial Neural Networks,
  Computational Intelligence and Machine Learning.

\bibitem[{Goodman(1999)}]{Goodman99}
Goodman, S.~N., 1999. Toward evidence-based medical statistics. 2: The {B}ayes
  factor. Annals of internal medicine 130~(12), 1005--1013.

\bibitem[{Herbrich et~al.(2001)Herbrich, Graepel, and Campbell}]{Herbrich01}
Herbrich, R., Graepel, T., Campbell, C., January 2001. Bayes point machines.
  Journal of Machine Learning Research 1, 245--279.

\bibitem[{Klein(2004)}]{klein2004sensor}
Klein, L.~A., 2004. Sensor and data fusion: a tool for information assessment
  and decision making. Vol. 324. Spie Press Bellingham, WA.

\bibitem[{Kschischang et~al.(2001)Kschischang, Frey, and
  Loeliger}]{kschischang2001factor}
Kschischang, F.~R., Frey, B.~J., Loeliger, H.-A., 2001. Factor graphs and the
  sum-product algorithm. IEEE Transactions on information theory 47~(2),
  498--519.

\bibitem[{Liu and Motoda(1998)}]{Liu98}
Liu, H., Motoda, H., 1998. Feature extraction, construction and selection: A
  data mining perspective. Springer.

\bibitem[{May et~al.(2012)May, Korda, Lee, and Leslie}]{may2012optimistic}
May, B.~C., Korda, N., Lee, A., Leslie, D.~S., Jun. 2012. Optimistic bayesian
  sampling in contextual-bandit problems. J. Mach. Learn. Res. 13, 2069--2106.

\bibitem[{Minka(2001)}]{minka2001expectation}
Minka, T.~P., 2001. Expectation propagation for approximate {B}ayesian
  inference. In: Proceedings of the Seventeenth conference on Uncertainty in
  artificial intelligence. Morgan Kaufmann Publishers Inc., pp. 362--369.

\bibitem[{Opper(1998)}]{Opper98}
Opper, M., 1998. A {B}ayesian approach to on-line learning. In: Saad, D. (Ed.),
  On-line Learning in Neural Networks. Cambridge University Press, New York,
  NY, USA, pp. 363--378.

\bibitem[{Rashidi and Cook(2011)}]{Rashidi11}
Rashidi, P., Cook, D.~J., Jun. 2011. Activity knowledge transfer in smart
  environments. Pervasive Mob. Comput. 7~(3), 331--343.

\bibitem[{Snoek et~al.(2005)Snoek, Worring, and Smeulders}]{snoek2005early}
Snoek, C.~G., Worring, M., Smeulders, A.~W., 2005. Early versus late fusion in
  semantic video analysis. In: Proceedings of the 13th annual ACM international
  conference on Multimedia. ACM, pp. 399--402.

\bibitem[{Twomey et~al.(2016{\natexlab{a}})Twomey, Diethe, and
  Flach}]{twomey2016structure}
Twomey, N., Diethe, T., Flach, P., 2016{\natexlab{a}}. On the need for
  structure modelling in sequence prediction. Machine Learning 104~(2),
  291--314.
\newline\urlprefix\url{http://dx.doi.org/10.1007/s10994-016-5571-y}

\bibitem[{Twomey et~al.(2016{\natexlab{b}})Twomey, Diethe, Kull, Song,
  Camplani, Hannuna, Fafoutis, Zhu, Woznowski, Flach, and
  Craddock}]{twomey2016sphere}
Twomey, N., Diethe, T., Kull, M., Song, H., Camplani, M., Hannuna, S.,
  Fafoutis, X., Zhu, N., Woznowski, P., Flach, P., Craddock, I.,
  2016{\natexlab{b}}. The {SPHERE} challenge: Activity recognition with
  multimodal sensor data. arXiv preprint arXiv:1603.00797.

\bibitem[{Winn et~al.(2015)Winn, Bishop, and Diethe}]{winn2015model}
Winn, J., Bishop, C.~M., Diethe, T., 2015. Model-Based Machine Learning.
  Microsoft Research Cambridge.
\newline\urlprefix\url{http://www.mbmlbook.com}

\bibitem[{Winn and Bishop(2005)}]{winn2005variational}
Winn, J.~M., Bishop, C.~M., 2005. Variational message passing. In: Journal of
  Machine Learning Research. pp. 661--694.

\bibitem[{Woznowski et~al.(2016)Woznowski, Burrows, Camplani, Diethe, Fafoutis,
  Hall, Hannuna, Kozlowski, Twomey, Tan, Zhu, Elsts, Vafeas, Mirmehdi,
  Burghardt, Damen, Paiement, Tao, Flach, Oikonomou, Piechocki, and
  Craddock.}]{woznowski2016sphere}
Woznowski, P., Burrows, A., Camplani, M., Diethe, T., Fafoutis, X., Hall, J.,
  Hannuna, S., Kozlowski, M., Twomey, N., Tan, B., Zhu, N., Elsts, A., Vafeas,
  A., Mirmehdi, M., Burghardt, T., Damen, D., Paiement, A., Tao, L., Flach, P.,
  Oikonomou, G., Piechocki, R., Craddock., I., 2016. {SPHERE}: {A} {S}ensor
  {P}latform for {HE}althcare in a {R}esidential {E}nvironment. In: Angelakis,
  V., Tragos, E., P{\"o}hls, H., Kapovits, A., Bassi, A. (Eds.), Designing and
  Developing and and Facilitating Smart Cities: Urban Design to {IoT}
  Solutions. Springer.

\bibitem[{Woznowski et~al.(2015)Woznowski, Fafoutis, Song, Hannuna, Camplani,
  Tao, Paiement, Mellios, Haghighi, Zhu, et~al.}]{Woznowski15}
Woznowski, P., Fafoutis, X., Song, T., Hannuna, S., Camplani, M., Tao, L.,
  Paiement, A., Mellios, E., Haghighi, M., Zhu, N., et~al., 2015. A multi-modal
  sensor infrastructure for healthcare in a residential environment. In:
  Communication Workshop {(ICCW)}, 2015 {IEEE} International Conference on.
  IEEE, pp. 271--277.

\bibitem[{Zhu et~al.(2015)Zhu, Diethe, Camplani, Tao, Burrows, Twomey, Kaleshi,
  Mirmehdi, Flach, and Craddock}]{zhu2015bridging}
Zhu, N., Diethe, T., Camplani, M., Tao, L., Burrows, A., Twomey, N., Kaleshi,
  D., Mirmehdi, M., Flach, P., Craddock, I., 2015. Bridging e-health and the
  internet of things: The {SPHERE} project. Intelligent Systems, IEEE 30~(4),
  39--46.

\end{thebibliography}

\end{document}